\documentclass{article}

\PassOptionsToPackage{numbers, compress}{natbib}

\usepackage[final]{neurips_2024}

\usepackage[utf8]{inputenc} 
\usepackage[T1]{fontenc}    
\usepackage{hyperref}       
\usepackage{url}            
\usepackage{booktabs}       
\usepackage{amsfonts}       
\usepackage{nicefrac}       
\usepackage{microtype}      
\usepackage{xcolor}         
\usepackage{graphicx}
\usepackage{wrapfig}
\usepackage{tcolorbox}
\usepackage{multirow}

\title{Evaluating Explanations Through LLMs: \\ Beyond Traditional User Studies}

\author{
\begin{tabular}[t]{c}\bfseries\rule{0pt}{24pt}
Francesco Bombassei De Bona \\
Università della Svizzera italiana \\
\texttt{francesco.bombassei.de.bona@usi.ch} \\
\end{tabular}\hfil\nolinebreak\hfil%
\begin{tabular}[t]{c}\bfseries\rule{0pt}{24pt}\ignorespaces%
Gabriele Dominici \\
Università della Svizzera italiana \\
\texttt{gabriele.dominici@usi.ch} \\
\end{tabular}\hfil\linebreak[4]\hfil%
\centerline{
\begin{tabular}[t]{c}\bfseries\rule{0pt}{24pt}\ignorespaces%
Tim Miller \\
The University of Queensland \\
\texttt{timothy.miller@uq.edu.au} \\
\end{tabular}\hfil\nolinebreak\hfil%
\begin{tabular}[t]{c}\bfseries\rule{0pt}{24pt}\ignorespaces%
Marc Langheinrich \\
Università della Svizzera italiana \\
\texttt{marc.langheinrich@usi.ch} \\
\end{tabular}\hfil\nolinebreak\hfil%
\begin{tabular}[t]{c}\bfseries\rule{0pt}{24pt}\ignorespaces%
Martin Gjoreski \\
Università della Svizzera italiana \\
\texttt{martin.gjoreski@usi.ch} \\
\end{tabular}%
}
}

\begin{document}
\maketitle

\begin{abstract}
As AI becomes fundamental in sectors like healthcare, explainable AI (XAI) tools are essential for trust and transparency. However, traditional user studies used to evaluate these tools are often costly, time consuming, and difficult to scale. In this paper, we explore the use of Large Language Models (LLMs) to replicate human participants to help streamline XAI evaluation.
We reproduce a user study comparing counterfactual and causal explanations, replicating human participants with seven LLMs under various settings. Our results show that (i) LLMs can replicate most conclusions from the original study, (ii) different LLMs yield varying levels of alignment in the results, and (iii) experimental factors such as LLM memory and output variability affect alignment with human responses.
These initial findings suggest that LLMs could provide a scalable and cost-effective way to simplify qualitative XAI evaluation.
\end{abstract}

\section{Introduction}
\label{intro}

As artificial intelligence (AI) becomes integrated into critical sectors such as healthcare \citep{rajpurkar_chexnet_2017, hannun_cardiologist-level_2019, rajkomar_scalable_2018}, the adoption of explainable AI (XAI) becomes inevitable \citep{sadeghi_brief_2023, saraswat_explainable_2022, mariappan_extensive_2024}. For example, AI models can help diagnose diseases \citep{chen_pan-cancer_2022}, predict patient outcomes \citep{hemker_healnet_2023}, and recommend treatments \citep{komorowski_artificial_2018}. The decisions of these models are often opaque, making it difficult for practitioners to fully trust or understand them. Therefore, XAI tools can have a huge impact in the integration of AI in healthcare. This necessity is also highlighted by regulatory efforts such as the EU AI Act \citep{panigutti_role_2023}, which enforces transparency and accountability in AI systems, particularly in critical sectors, where understanding AI-driven decisions can mean the difference between life and death.

This need for effective XAI tools has led to a significant number of studies aimed at advancing the field \citep{barredo_arrieta_explainable_2020}. Many of these efforts have focused mainly on developing new techniques and algorithms \citep{ribeiro_why_2016, lundberg_unified_2017, selvaraju_grad-cam_2020, koh_concept_2020, kim_interpretability_2018, wachter_counterfactual_2018} to explain models and evaluate them through quantitative metrics. However, this approach holds significant challenges, as there are no clear and unique metrics (e.g., surrugate model fidelity \citep{ribeiro_why_2016}, counterfactual validity \citep{wachter_counterfactual_2018}, and proximity score \citep{guyomard_vcnet_2022}) to evaluate these tools. The choice of metrics is often highly dependent on the specific XAI technique and the domain of application, and these metrics frequently fail to capture the actual benefits from the end-user’s perspective. As a result, many tools are optimized to maximize performance in these quantitative metrics, ignoring the ultimate goal of providing explanations that help users understand the model’s decisions \citep{keane_if_2021}. In contrast, fewer studies involve qualitative evaluations in which users assess key properties such as the effectiveness, helpfulness, and trustworthiness of the explanations \citep{colin_what_2023, singh_actionability_2024, celar_how_2023, karagoz_evaluating_2024, rong_towards_2023}. Furthermore, there is no standardized process for structuring these evaluations, leading to inconsistencies in the way user studies are conducted. Consequently, running user studies tends to be not only costly and time consuming but also prone to producing variable outcomes, which limits their scalability and reproducibility. These challenges create barriers that result in fewer qualitative evaluations and slow down progress in the field.
 
Under these circumstances, Large Language Models (LLMs) offer a promising way to complement user studies. First, LLMs can help researchers run smaller, more focused studies by integrating their results with LLM-generated data, reducing the need for large-scale participant recruitment. This streamlines the evaluation of XAI tools while ensuring alignment with human preferences.
Second, LLMs are useful in expert-driven studies, where recruiting participants like clinicians is challenging. Instead of relying on large groups of laypeople for early-stage feedback—which can reduce the validity of the study—LLMs can provide preliminary insights, allowing researchers to refine tool designs before engaging experts, saving time and resources. For example, a healthcare institution developing machine learning models to detect brain cancer via MRI scans must validate the model by understanding its decision-making process. XAI techniques are essential for this, as they provide insights into the model’s reasoning. To ensure the model’s explanations align with clinical expectations, a user study involving practitioners is crucial. In this context, LLMs can streamline the process by pre-evaluating the XAI outputs, ensuring that the model’s decisions are coherent before full expert validation, saving time and resources.

This paper explores the potential of LLMs to bridge the gap between the need for scalable XAI evaluation and the limitations of traditional user studies. Specifically, we aim to replicate a user study which compared counterfactual and causal explanations in terms of their helpfulness and effectiveness in transmitting insights from AI systems. However, instead of human participants, we use LLMs and explore whether the LLM-generated results align with the conclusions drawn from the original user study.
We evaluated seven of the most advanced LLMs — Llama 3 (8B and 70B) \citep{touvron_llama_2023}, Qwen 2 (7B and 72B) \citep{yang_qwen2_2024}, Mistral 7B \citep{jiang_mistral_2023}, Mistral Nemo and GPT-4o Mini \citep{openai2024gpt4} - in various experimental settings. These settings included leveraging LLMs memory and exploring the effects of LLM variability in generating answers to understand their impact in the alignment between LLMs and humans preferences. 
The results of our experiments demonstrate that: (i) LLMs can replicate most of the conclusions from the original user study, (ii) different LLMs can lead to varying conclusions depending on the architecture and capabilities of the model, and (iii) the experimental setup, such as the use of memory or randomness, can significantly impact the extent to which LLM responses align with human responses. These initial findings provide promising insights into the feasibility of developing automatic, scalable, and cost-effective qualitative evaluation frameworks that rely on LLMs as an alternative to traditional user studies.

\begin{figure}
    \centering
    \includegraphics[width=0.65\linewidth, trim={0, 0, 0, 2cm}]{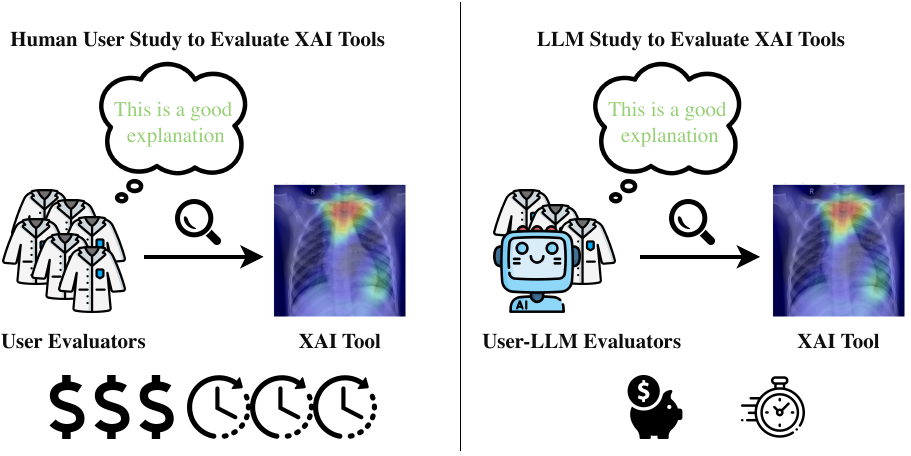}
    \caption{Our vision in comparing Human and LLM Evaluators for XAI Tool Effectiveness}
    \label{fig:human-vs-llm}
\end{figure}

\section{Traditional user studies to evaluate XAI tools}
\label{user-study}

In the evaluation of XAI tools, particularly within healthcare, user studies are considered the gold standard for assessing how well these tools perform in real-world scenarios \citep{doshi-velez_towards_2017}. Typically, these studies involve a structured process in which healthcare professionals or end users interact with XAI tools under controlled conditions. In previous user studies that evaluated XAI tools \citep{colin_what_2023, singh_actionability_2024, celar_how_2023, karagoz_evaluating_2024}, participants were assigned with activities such as predicting the output of the AI model based on AI-generated explanations and/or completing questionnaires about the perceived usefulness of the explanations and the degree of trust in the model.

For instance, \citet{colin_what_2023} asked participants to guess the prediction of the model based on the provided explanation to evaluate how various XAI techniques \citep{selvaraju_grad-cam_2020, simonyan_deep_2014} help users detect biases, identify strategies for solving unknown tasks, and recognize model failures. Their user study included scenarios with inherent biases, tasks in which users lacked domain expertise, and models prone to misclassifying specific examples. The primary objective was to assess which explanation strategies were most effective in helping users replicate the model’s decision-making process. The assumption was that if an explanation was understandable and complete, users would be able to predict the model’s decisions accurately. The conclusions were drawn by comparing the performance of different explanation techniques against each other, as well as against a baseline scenario where no explanations were provided.
Similarly, \citet{karagoz_evaluating_2024} asked professionals to report their level of trust in an AI model both before and after receiving an explanation \citep{ribeiro_why_2016, lundberg_unified_2017} of the decision of the model. Participants were also asked to make predictions based on the explanation. Additionally, the study examined the level of agreement between practitioners who made decisions using XAI tools versus those who did not. As with the previous study, the researchers used statistical tests to compare the results across different explanations and settings (pre- and post-explanation).
In a different approach, \citet{singh_actionability_2024} designed a study in which participants filled out a questionnaire with several items regarding the actionability of the explanations they were shown. The aggregated results were compared to determine which types of explanations were perceived as more actionable. Their findings demonstrated that this setting effectively highlighted which explanation type was considered most actionable by participants.
These user studies demonstrate the diversity of metrics and settings used (from performing tasks to giving opinion through a questionnaire) in the evaluation of XAI tools, highlighting the lack of a universally accepted structure or standardized metrics for conducting user studies in this field. Despite this variability, one common thread across all the studies is the comparison of aggregated results for different XAI tools in hypothetical real-world scenarios. This ability to simulate real-world use cases is the primary advantage of user studies.

However, despite their importance, user studies have several significant drawbacks. Conducting these studies is often resource intensive and requires substantial time and financial investment to recruit participants and simulate usage environments (e.g., clinical decision making in the context of healthcare). The involvement of domain experts (e.g., clinicians), who may already have demanding schedules, further complicates the process. Furthermore, variability in user experience and interpretation can lead to inconsistent results, making it challenging to draw generalizable conclusions, especially if the number of participants is relatively low. This variability also limits the scalability of user studies, as replication across different institutions, or with larger groups, can be prohibitively expensive and time consuming.

\section{Can LLMs evaluate XAI tools?}
\label{llm-user-study}

LLMs are advanced AI systems designed and trained to process text data and generate human-like text based on vast amounts of data, covering a wide range of topics. This training enables them to estimate context, generate coherent answers, and mimic human-like reasoning in their responses. LLMs excel in tasks that require the comprehension and generation of natural language, making them particularly effective in simulating human interactions \citep{openai_gpt-4_2024}, such as those involved in user studies.
LLMs have also demonstrated significant performance in tasks beyond their original training without the need for additional fine-tuning. For example, LLMs can classify various data types, such as tabular data \citep{hegselmann_tabllm_2023} and time series \citep{gruver_large_2024}, or generate synthetic data \citep{li_synthetic_2023}. Although they may not yet represent the state-of-the-art models for these tasks (with specialized models often being more capable), they offer valuable versatility. This is especially relevant when dealing with niche domains like healthcare, where specialized models are typically preferred for specific tasks due to their superior and tailored capabilities. 
However, LLMs can offer additional capabilities in conjunction with specialized models and tools, leveraging their human-like conversational abilities and contextual understanding.

To exploit these abilities in conjunction with specialized models and XAI tools, we propose using LLMs to qualitatively evaluate XAI tools, simulating human-based studies. Querying LLMs instead of humans offers several significant advantages. Firstly, LLMs provide a cost-effective alternative, eliminating the need for expensive and time-consuming recruitment and compensation of human participants. This allows for the rapid and inexpensive execution of studies that can be conducted repeatedly with unlimited queries, leading to large-scale data collection and analysis.
In general, including the use of LLMs in the user study would dramatically increase the flexibility and scalability of the research process regarding the development of XAI tools. 

However, achieving useful results with LLMs is only possible if the LLM model is properly aligned with human preferences. Ensuring such alignment is one of the key challenges in the development of more powerful LLMs today. This alignment is typically achieved through additional training \citep{ouyang_training_2022} or techniques \citep{irving_ai_2018}. These differences in the alignment process can significantly affect the way LLMs generate responses, influencing their reasoning processes, beliefs, and preferences.
This variability in alignment becomes a critical consideration in scenarios where the primary goal is to accurately mimic user preferences, such as in our evaluations of XAI tools. In particular, we are interested in assessing two types of alignment: \textit{general alignment}, where we determine the outcome of statistical comparisons (e.g., explanation A is generally perceived as better than explanation B), and \textit{absolute alignment}, where we measure specific ratings (e.g., explanation A is rated a 4 on a scale of 1 to 5 for helpfulness).
In both cases, the choice of which LLM to use is not trivial. Different LLMs can exhibit varying degrees of alignment depending on the training process and the intended application. Thus, selecting the most suitable LLM is a critical design decision that can significantly impact the effectiveness and reliability of the user study.

In addition to alignment, several other factors influence how LLMs generate responses. These include varying the initial prompts, employing different prompting techniques (e.g., zero-shot, few-shot \citep{brown_language_2020}, chain-of-thought \citep{wei_chain--thought_2023}) and leveraging session memory. The structure and clarity of the input prompt can significantly affect the coherence and relevance of the model output. For example, various types of prompt injection can be used to alter the generation process, incorporating elements such as personalization, task description, and context specification. Then, conversation memory can impact LLMs' performances by allowing the models to exploit prior generated information, increasing the chance of alignment (as also the human evaluator is influenced by the past examples) and also modifying the variability of the output. 
If memory is used, then the order of instruction also plays an important role in conditioning the results.
These factors contribute to the variability in LLM generation, further highlighting the importance of carefully selecting and configuring the model to better align with specific user preferences and study goals.

\section{Experiments}
\label{sec:exp}
We designed our experiments to explore whether it is feasible to estimate human preferences in XAI user studies using LLMs.  More specifically, we tested for \textit{general alignment} (i.e., can we arrive to the same findings/conclusions on a population/study level) and \textit{absolute alignment} (i.e., can we come to similar responses on a case-by-case level) between our LLM-based study and an existing use-based study. Additionally, we aim to explore the impact of different factors that might influence the outcomes of such studies when using LLMs. To achieve this, we address the following research questions:

\begin{itemize}
\item \textbf{Alignment}: Is it possible to replicate the results of an XAI user study using LLMs? Are LLMs answers aligned with human preferences in the context of evaluating explanations in absolute terms? 
\item \textbf{LLM Variability}: Do different LLMs lead to different general alignment? 
\item \textbf{Framework Variability}: Does the use of memory in LLMs influence the results of the user study? How does variability in LLM responses impact the general alignment?
\end{itemize}

\subsection{Evaluation Setting}
\label{sec:eval}

\subsubsection{User study}

As a foundation for the experimental setting, we use the first set of experiments of the user study by \citet{celar_how_2023}. Their study was designed to better understand the relationship between causal and counterfactual explanations in terms of their helpfulness and effectiveness in transmitting insights from AI systems to end users.
Participants interacted with the predictive AI system's input and output, along with explanations from an XAI tool providing insights into the system's decision-making process. The study featured four experimental conditions: \textit{counterfactual }versus \textit{causal} explanations in \textit{high} versus \textit{low familiarity} scenarios. \textit{Counterfactual} explanations provide alternative scenarios by answering "what if" questions and describing how the world would have to change for a desirable outcome to occur. In contrast, \textit{causal explanations} describe the direct cause-and-effect relationships that led to the observed outcome. In the \textit{high familiarity} condition, participants determine whether alcohol levels are over or under the legal limit for driving. In contrast, the \textit{low familiarity} condition requires participants to assess the safety of an unknown chemical compound.

In each experimental setting of their study, users completed three tasks. In the first, they rated the helpfulness of a given explanation based on the input and output of the AI system. In the second task, participants attempted to make the prediction themselves using the provided explanation. Finally, in the third, they expressed their confidence level in their prediction on the Likert scale: "not at all confident", "not very confident", "neither", "fairly confident", and "very confident".

The first two tasks consisted of sixteen cases each. In the \textit{high familiarity} scenario, the case comprised the following fields: name of the subject, weight, units of alcohol consumed, duration, gender, and stomach content. In the \textit{low familiarity} setting, the case comprised the name of the chemical, occupational exposure limit, pH, exposure duration, air pollution rating, and PNEC Rating.
Participants completed the first task on sixteen cases, followed by the second task on sixteen new cases. The third task questions were interleaved between the second task cases in equal numbers, ensuring sequential progression.
In the first task, each case was paired with a prediction and an explanation, either causal or counterfactual, and the user judged the statement "This explanation was helpful" on a Likert scale: "strongly disagree", "disagree", "neutral", "agree", and "strongly agree".
In the second task, only the case information was shown to enable the user to make their prediction, either over the limit/under the limit or safe/not safe.

\subsubsection{Estimating human preferences with LLMs}

To replicate the above-mentioned user study \citep{celar_how_2023}, we transpose the experimental setting designed for human evaluators into a compatible setting for LLMs. In this context, each LLM is treated as a participant, tasked with generating responses across the same experimental conditions (high/low familiarity and causal/counterfactual explanations).

LLM models are used to generate responses in place of human participants. A run corresponds to the execution of one experiment formed by the helpfulness, prediction, and confidence tasks. A state refers to the specific conditions under which the model generates responses, such as the combination of familiarity and explanation type for each task.
We use two approaches to aggregate and compare the results across different models and runs. In the first approach, we conduct multiple inference runs for each model and calculate the mean response at each question. This method, similar to the self-consistency technique, helps to mitigate the variability in the generated outputs and produces a more stable, reliable average response for each experimental condition. By averaging across multiple runs, we reduce the impact of any outlier responses and ensure consistency in the model’s performance.
In the second approach, we treat each inference run as if it were generated by a different participant. This method simulates the diversity typically seen in a group of human participants. By treating each run independently, we capture the variability that may arise in model-generated responses, mirroring how individual differences exist in human participants.

In addition, the use of instruction-following models enables us to explore the influence of conversation history during task execution. We assume that while performing the tasks, users undergo a learning process. Thus, to replicate the same process, we enable the LLM to use previously generated answers and instructions as context for every new inference. We test this scenario against a baseline where the models do not have access to any previously generated answers and inputs, and every inference is treated as a separate task. We refer to these two scenarios as \textit{"with memory"} and \textit{"in isolation"}, respectively.
The isolation setting poses two challenges. The first issue arises in the third task, where the LLM is asked to provide a confidence level about the prediction made in the previous task. However, if each inference is treated without memory, the LLM has no information about the previous task or the answer given. As a result, it cannot express a confidence level about its performance. To address this issue, we switch to a hybrid approach regarding memory by allowing the second and third tasks to be completed in pairs and enabling the conversation history between the two tasks.
The second issue involves the assumption that the users, while performing the first task learn how to make the predictions in the second task. This assumption is violated due to the lack of presence of the conversation memory.
To address this issue we apply a few-shot prompting technique. This method uses synthetic data of input-output examples to instruct the LLM on what answer is expected and the way reasoning should be done. In our scenario, we take the sixteen cases from the first task, pair them with the corresponding prediction as if made by the same LLM, and use them as context for the input of the second task.

Given the two initial settings (familiar and not familiar) for our experiments, we feed LLMs with specific prompts to fit each task and each conversation memory setting.
Specifically, we propose the following prompts, where we switch the familiarity setting based on the experiment and inject the cases illustrated previously:

\begin{tcolorbox}[colback=green!5!white,colframe=green!75!black,title=First task prompt, fontupper=\footnotesize, fonttitle=\footnotesize, enlarge bottom by=-2mm, enlarge top by=-2mm, boxsep=0mm]
    Given the following case, how would you rate the sentence "This explanation was helpful"? You must answer by only providing one value from the following: "Strongly disagree", "Disagree", "Neutral", "Agree", "Strongly agree". \{case\}
\end{tcolorbox}

\begin{tcolorbox}[colback=green!5!white,colframe=green!75!black,title=Second task prompt, fontupper=\footnotesize, enlarge bottom by=-2mm, fonttitle=\footnotesize, boxsep=0mm]\label{tcolorbox:second}
    Complete the sentence "Based on the information provided, I believe the app's prediction for this person|chemical will be ...". You must answer by only providing one value from the following: Over the limit, Under the limit | Safe, Not safe. \{case\}
\end{tcolorbox}

\begin{tcolorbox}[colback=green!5!white,colframe=green!75!black,title=Third task prompt, fontupper=\footnotesize, fonttitle=\footnotesize, fonttitle=\footnotesize, enlarge bottom by=-1mm, boxsep=0mm]\label{tcolorbox:third}
    How confident are you in your prediction? You must answer by only providing one value from the following: "Not at all confident", "Not very confident", "Neither", "Fairly confident", "Very confident".
\end{tcolorbox}

Our final consideration concerns the impact of case ordering during inference in the memory setting. Since the original user study used different permutations of the cases for each participant, we apply the same approach to the LLM to ensure consistency. This allows us to account for any potential influence that case order might have on performance or results.
This introduces the concept of “LLM-user.” An LLM-user refers to the aggregation of results obtained across multiple inference runs, where a fixed permutation of cases is used for each group of runs. The results from multiple LLM-users are then combined to form the study's overall conclusions.

\subsubsection{Metrics and statistical tests}

Our primary objective is to compare the results obtained from LLMs with those from the original user study. To ensure consistency, we use the same statistical metrics and tests as those proposed in the original user study \citep{celar_how_2023}.

The original study compares four experimental conditions: low familiarity with causal explanations, low familiarity with counterfactual explanations, high familiarity with causal explanations, and high familiarity with counterfactual explanations. The analysis focuses on the mean responses provided by participants, aiming to show that high familiarity scenarios lead to better outcomes compared to low familiarity ones and that counterfactual explanations are generally more helpful and insightful than causal explanations. We replicate this approach by calculating the mean values for the LLM-generated responses in each condition and comparing the outcomes across the four experimental scenarios. This allows us to get information about the absolute alignment between LLMs and humans. 

However, the main conclusion of the paper are drawn on statistical tests regarding the effect of the two primary variables — familiarity (high vs. low) and explanation type (causal vs. counterfactual) — and their interaction. Therefore, we apply a two-way ANOVA statistical test \citep{sthle_analysis_1989}, just as in the original study, to assess whether these factors significantly influence the responses of LLMs, as they do for human participants (general alignment). This method allows us to investigate whether LLMs exhibit similar patterns of reasoning and judgment, even if they do not present an absolute alignment.
Since we aim to assess the alignment between LLMs and the user study, we assume that the answer distributions match in shape and, therefore, we test for normality of the LLMs’ answers. The ANOVA test assumptions are only partially satisfied for certain models, specifically Mistral-Nemo-Instruct-2407, Mistral-7B-Instruct-v0.3, and GPT-4o-Mini. As a result, we exercise caution and refrain from drawing strong conclusions based on the statistical outcomes for these models.

By aligning our evaluation techniques with those used in the original work, we aim to determine how well LLMs replicate human reasoning processes and whether they demonstrate the same preferences and patterns when presented with familiar or unfamiliar scenarios, as well as causal or counterfactual explanations.

\subsection{Results}

\begin{wrapfigure}{r}{0.57\textwidth}
    \centering
    \includegraphics[width=0.57\textwidth, trim={0, 2cm, 0, 2cm}]{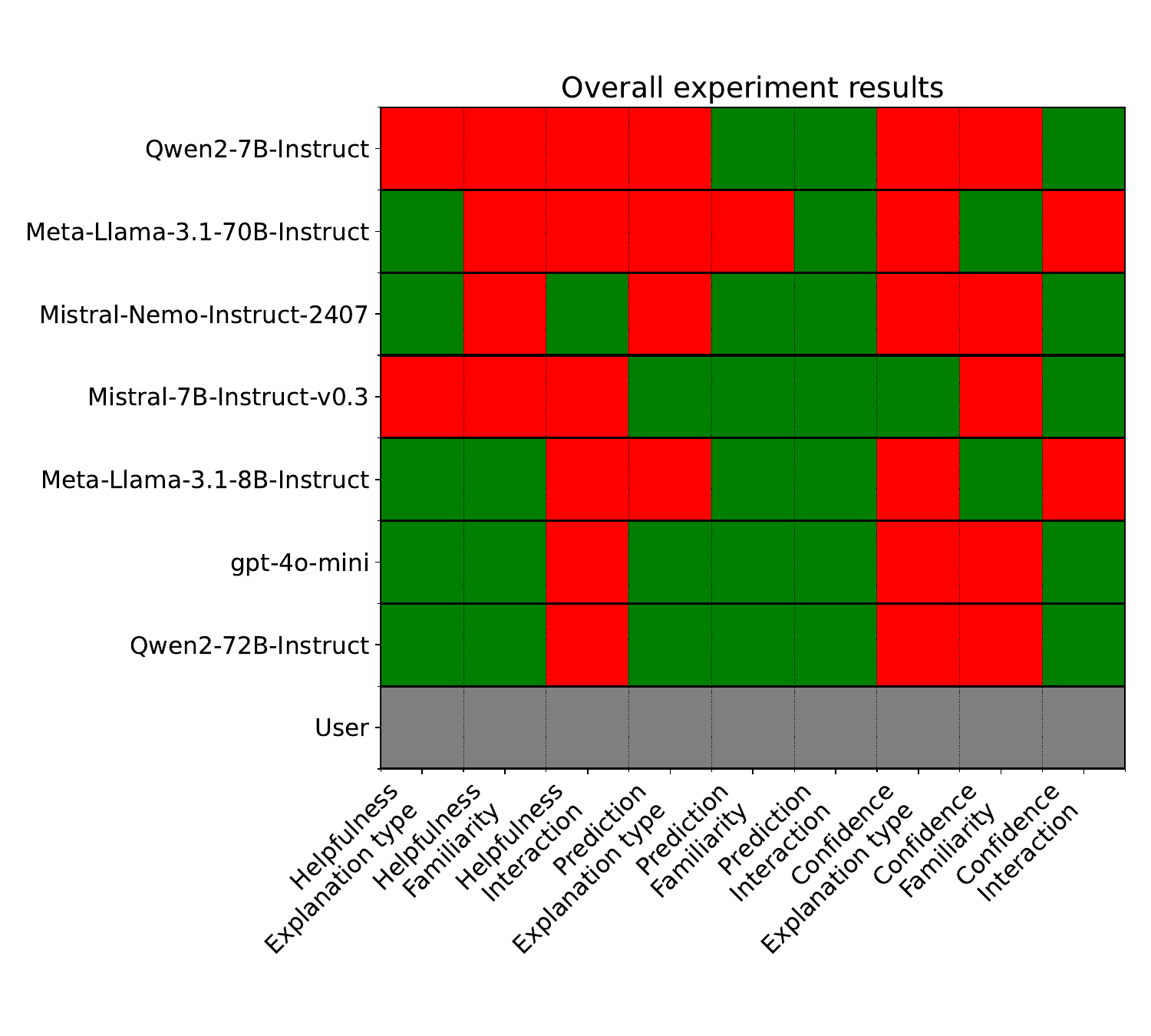}
    \caption{Concordance of the results for each statistical test. Concordance is computed by merging the results of the ANOVA test and the results of the comparison of the averaged values.}
    \label{fig:overall-heatmap}
\end{wrapfigure}

\paragraph{Evaluating XAI tools with LLMs partially mirrors user study conclusions. (Figure \ref{fig:overall-heatmap})}
Using the results of the Qwen 2 72B model (sampling the same number of “LLM-users” as participants in the original user study), we successfully replicated 6 of the 9 statistical outcomes from the first part of the user study. Figure \ref{fig:overall-heatmap} illustrates the concordance between the LLM results and those of the original human-based study for each statistical test. This shows that, under specific conditions, partial alignment between human and LLM conclusions can be achieved when replacing human participants with LLMs. Specifically, we observed perfect general alignment in tasks that require the prediction of the model output given an explanation. However, alignment was more challenging in tasks that involved confidence-related questions, where the LLM struggled to match human responses.

\begin{figure}[t]
    \centering
    \includegraphics[width=0.8\linewidth]{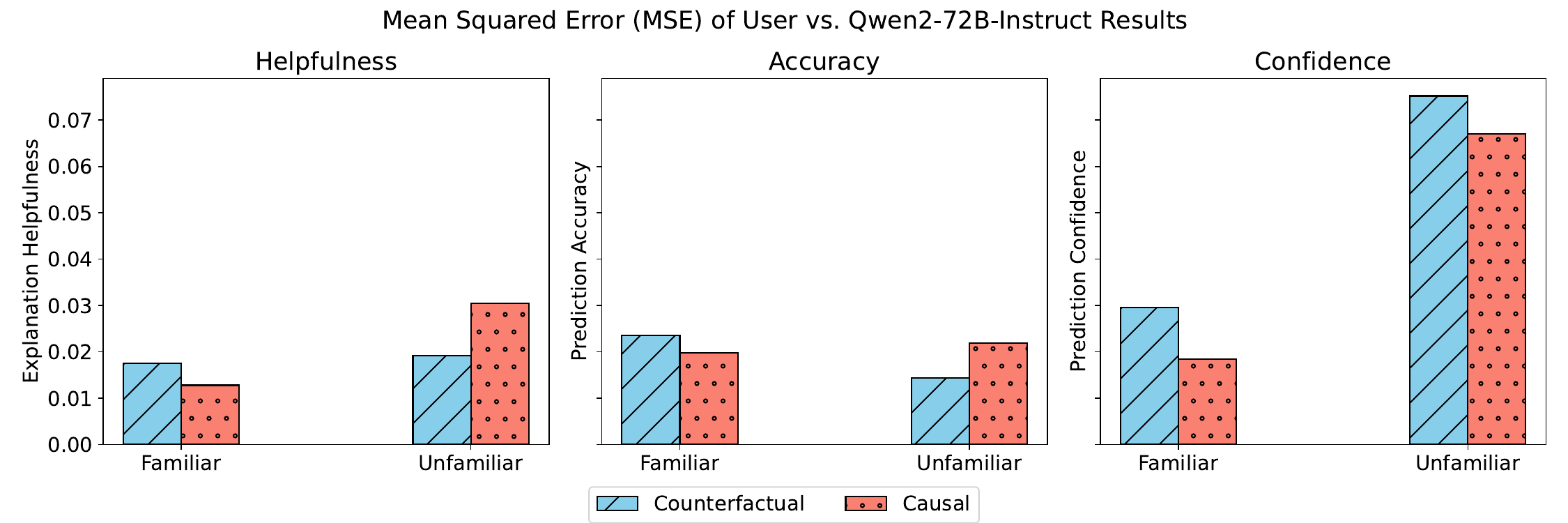}
    \caption{Bar charts representing the MSE between users and Qwen2-72B in the three tasks}
    \label{fig:overall-mse-qwen}
\end{figure}

\paragraph{LLMs exhibit slight differences in absolute alignment with human preferences. (Figure \ref{fig:overall-mse-qwen})}

Although LLMs can replicate overall trends in user studies, their responses still show some deviations from human participants. Figure \ref{fig:overall-mse-qwen} presents the MSE of the Qwen 2 72B model across different categories (helpfulness, accuracy, confidence) in both familiar and unfamiliar conditions under causal and counterfactual settings. While the MSE for helpfulness is relatively low, particularly in familiar contexts, the model struggles more with accuracy, especially in unfamiliar settings. Confidence shows the largest errors, mainly in unfamiliar conditions. These results suggest that, although the model’s absolute predictions differ from human results, its comparative judgments remain consistent, showing potential as an alternative in user evaluations across different scenarios.

\paragraph{Different LLMs exhibit varying levels of general alignment with human responses. (Figure \ref{fig:overall-heatmap})}
The degree of general alignment between LLMs and human participants varies between models, reflecting differences in size, capabilities, and behaviors. Figure \ref{fig:overall-heatmap} shows that some LLMs align more closely with human judgments in specific tasks, while others diverge. Among the tested models, Qwen 2 72B and GPT-4o Mini achieved the highest general alignment, matching human conclusions in 6 out of 9 cases. Interestingly, the smaller LLaMA 8B demonstrated better general alignment than its larger counterpart, LLaMA 70B. In contrast, the largest Qwen 2 model (72B) significantly outperformed the smaller Qwen 2 7B, indicating that a larger model does not necessarily guarantee better alignment with human responses across all architectures.
Additionally, all models showed their best performance on the prediction tasks, while they performed worst in the confidence-related tasks. This may be because LLMs are trained to answer specific questions (such as predicting outcomes) based on provided context (an explanation). Confidence questions, on the other hand, involve subjective judgments that are more challenging for LLMs to mimic, as they may struggle to express confidence levels—since their responses always represent the most probable outcome, not a measure of certainty. This suggests that while LLMs can replicate decision-making tasks, more nuanced, subjective metrics like confidence may require further refinement.

\begin{figure}[t]
    \centering    \includegraphics[width=.95\linewidth, trim={0, 2cm, 0, 2cm}]{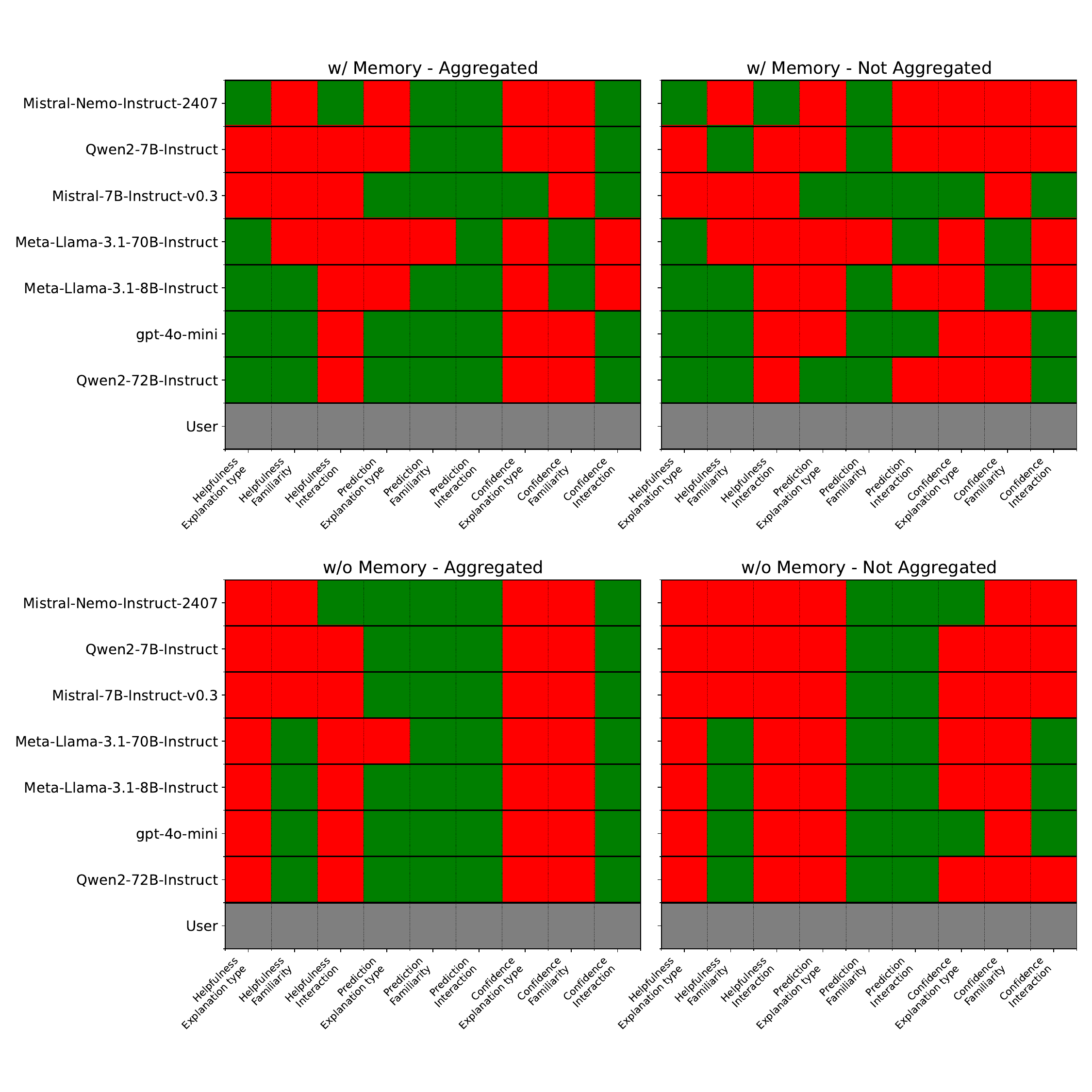}
    \caption{Concordance of the results for each statistical test aggregated by experimental settings. Experimental settings explored comprise conversation with or without conversation memory and usage of aggregated inference runs.}
    \label{fig:factor-comparison}
\end{figure}

\paragraph{Usage of memory impacts LLMs’ general alignment. (Figure \ref{fig:factor-comparison})}
Figure \ref{fig:factor-comparison} illustrates the concordance between the results of the LLMs and those of the original human-based study in different settings, specifically comparing models with and without memory use. LLMs that utilize memory behave differently from those that do not. Generally, LLMs without memory exhibit more uniform performance across all tests, likely due to the absence of influence from the prior context. In contrast, the use of memory introduces variability, as the model’s responses are affected by the way it interprets and incorporates information from previous interactions.
Despite this variability, models that employed memory tended to perform better overall, showing higher concordance with human judgments. This suggests that memory, when used effectively, can enhance an LLM’s ability to align with human responses. However, this benefit depends on how well the memory mechanism is utilized.

\paragraph{Simulating different users through aggregation leads to opposite results. (Figure \ref{fig:factor-comparison})} Evaluating the impact of aggregation allows us to better assess the impact of LLM generation variability. Figure \ref{fig:factor-comparison} compares the outcomes of different models using both aggregated and non-aggregated methods.
The results reveal that prior to aggregation, different LLMs exhibit varying degrees of initial variability, as they show different level of agreement with the aggregated results. Therefore, the aggregation method significantly influences the alignment of the LLM results. For instance, models with memory that use aggregation closely mirror the human user results.
However, when non-aggregated responses are considered, the variability across runs increases, leading to more divergent results, particularly in confidence ratings and prediction accuracy.
The non-aggregated approach without memory produces results that deviate significantly from the original user study. This suggests that relying on individual, un-aggregated LLM responses introduces greater variability, making it difficult to replicate human users. By contrast, aggregated models, especially those using memory, maintain more consistent performance. As shown in Figure \ref{fig:factor-comparison}, by using aggregation, we observe an increase in the alignment of LLMs with the user study in 11 out of 14 cases while the other 3 maintain the original alignment with the non-aggregated models.

\section{Limitations}
\label{sec:limitations}

One significant limitation of our research is that it is based on a single publicly available user study, which focused on evaluating explanations generated by two similar XAI methods. The use of only two explanation techniques in a single user study limits the breadth of the conclusions we can draw. XAI encompasses a wide variety of techniques and applications across different domains, and our findings may not generalize to all types of explanations or contexts.
Moreover, our approach cannot replicate certain types of human studies, such as qualitative interviews or assessments that measure real trust, particularly in expert-driven fields like healthcare, where the trust of clinicians is crucial. Similarly, this approach would struggle to handle entirely new domains that fall outside the LLM’s training set, as the models rely on prior knowledge to generate responses.
However, this study highlights an interesting path for further exploration. We plan to extend the experimental settings in future work to include a broader range of XAI techniques and user studies. Additionally, we aim to incorporate queries to Vision Language Models (VLMs) to evaluate visually oriented XAI techniques, such as saliency maps, which are important in the healthcare field.

Another limitation is the possibility that the tasks or responses from the original \citet{celar_how_2023} study may be part of the LLM training set. This could introduce bias and compromise the validity of the results. Ensuring that the LLM responses are genuinely independent of the study’s prior knowledge will be critical in addressing this limitation.

Lastly, a limitation lies in the specific LLMs used in this study. Although we used up-to-date LLMs at the time of evaluation, the rapid pace of advancements in AI technology, especially in LLMs, means that future models may exhibit different behavior, reasoning abilities, or alignment capabilities. Similarly, improvements in alignment techniques could lead to better alignment with human preferences, potentially altering the conclusions drawn in this study. As a result, future research will need to inspect LLM performance again as new models and alignment methods emerge. Nonetheless, this paper serves as an illustration of this idea, providing promising results and laying the groundwork for further investigation.

\section{Conclusions}
\label{sec:conclusions} 
In conclusion, our investigation into the use of Large Language Models (LLMs) to complement and integrate user studies for evaluating Explainable AI (XAI) tools offers promising initial results. By replicating a user study on counterfactual and causal explanations, we found: (i) LLMs can replicate most of the conclusions derived from traditional user studies, indicating their potential as scalable and cost-effective alternatives, (ii) different LLM architectures and capabilities can produce varying outcomes, emphasizing the importance of selecting appropriate models for specific evaluation tasks, and (iii) experimental factors, such as the use of memory and the impact of variability in generating responses, significantly affect the alignment between LLM and human preferences.
Our findings suggest LLM-based evaluations could greatly improve the scalability and reproducibility of XAI assessments. Future work should aim to enhance the alignment of LLMs with human judgment in the evaluation of XAI tools and explore the broader applicability of this approach across various XAI techniques and domains.

\begin{ack}
This study was funded by the Swiss National Science Foundation, through the projects XAI-PAC (PZ00P2\_216405) and TRUST-ME (205121L\_214991).
\end{ack}

\bibliographystyle{unsrtnat}
\bibliography{references}

\newpage
\appendix

\section{Implementation details}
\label{appendix:arch}

\paragraph{Running local and remote models}
We executed Hugging Face models (Mistral, Llama 3.1, and Qwen2 families) on a local server equipped with the following hardware:

\begin{itemize}
\item \textbf{CPU}: 2 x AMD EPYC 7513 32-Core Processor
\item \textbf{RAM}: 512 GB
\item \textbf{GPU}: 4 x RTX A6000 (48 GB VRAM each)
\end{itemize}

CUDA acceleration was utilized to parallelize and distribute computation across the GPUs, significantly speeding up the processing.

The total inference time for running the full experiment was approximately 76 hours.

For GPT 4o Mini, inference was run using the OpenAI API. Inference time depends on the usage Tier available on the API.

\paragraph{Code and licenses}
Our code implementation is built using Python 3.12 and leverages the open-source library LangChain \citep{chase_langchain_2022} (MIT License) to develop the inference infrastructure for both local and remote executions. All plots were generated using the Matplotlib \citep{Hunter:2007} (BSD License) library. The dataset used in the experiment is freely available, following the guidelines provided in the original paper \citep{celar_how_2023}.

\end{document}